\documentclass[runningheads]{llncs}
\usepackage{graphicx}

\usepackage{tikz}
\usepackage{comment}
\usepackage{amsmath,amssymb} 
\usepackage{color}

\usepackage[accsupp]{axessibility}
\usepackage[colorlinks=true]{hyperref}

\usepackage{booktabs}
\usepackage{epsfig}
\usepackage{multirow}

\makeatletter
\DeclareRobustCommand\onedot{\futurelet\@let@token\@onedot}
\def\@onedot{\ifx\@let@token.\else.\null\fi\xspace}

\makeatother

\usepackage{orcidlink}

\begin{document}
\pagestyle{headings}
\mainmatter
\def\ECCVSubNumber{6191}  

\title{ObjectBox: From Centers to Boxes for Anchor-Free Object Detection} 

\titlerunning{ObjectBox: From Centers to Boxes for Anchor-Free Object Detection}
%
\author{Mohsen Zand\orcidlink{0000-0001-8177-6000
} \and
Ali Etemad\orcidlink{0000-0001-7128-0220} \and
Michael Greenspan\orcidlink{0000-0001-6054-8770}}
\authorrunning{M. Zand et al.}
%
\institute{Dept. of Electrical and Computer Engineering, 
Ingenuity Labs Research Institute \\ Queen's University, Kingston, Ontario, Canada }
\maketitle

\begin{abstract}
We present ObjectBox, a novel single-stage anchor-free and
highly generalizable object detection approach. As opposed to both existing anchor-based and anchor-free detectors, which 
are more biased toward specific object scales
in their label assignments, we use only object center locations as positive samples and treat all objects equally in different feature levels regardless of the objects' sizes or shapes. Specifically, our label assignment strategy considers the object center locations as shape- and size-agnostic anchors in an anchor-free fashion, and allows learning to occur at all scales for every object. 
To support this, we define new regression targets as the distances from two corners of the center cell location to the four sides of the bounding box. Moreover, to handle scale-variant objects, we propose a tailored IoU loss to 
deal with boxes with different sizes.
As a result, our proposed object detector does not need any dataset-dependent hyperparameters to be tuned across datasets. We evaluate our method on MS-COCO 2017 and PASCAL VOC 2012 datasets, and compare our results to state-of-the-art methods. We observe that ObjectBox performs favorably in comparison to prior works. Furthermore, we perform rigorous ablation experiments to evaluate different components of our method.
Our code is available at:  \href{https://github.com/MohsenZand/ObjectBox}{https://github.com/MohsenZand/ObjectBox}

\keywords{Object detection, Anchor-free, Object center, MS-COCO 2017, PASCAL VOC 2012}
\end{abstract}

\section{Introduction}

Current state-of-the-art object detection methods, regardless of whether they are a two-stage~\cite{girshick2015fast}, \cite{he2017mask}, \cite{cai2018cascade} or a one-stage method~\cite{redmon2018yolov3}, \cite{zhang2019freeanchor}, \cite{tian2019fcos}, hypothesize bounding boxes, extract features for each box, and label the object class. They both conduct bounding box localization and classification tasks on the shared local features. A common strategy is to use hand-crafted dense anchors on convolutional feature maps to generate rich candidates for shared local features~\cite{ke2020multiple}, \cite{uzkent2020efficient}. These anchors generate a consistent distribution of bounding box sizes and aspect ratios, which are assigned based on the Intersection over Union (IoU) between objects and anchors. 

Object detection has been dominated by anchor-based methods~\cite{lin2017focal}, \cite{redmon2018yolov3} due to their great success. They however suffer from a number of common and serious drawbacks. First, using predefined anchors introduces additional hyperparameters to specify their sizes and aspect ratios,  which impairs generalization to other datasets. Second, anchors must densely cover the image to maximize the recall rate. A small number of anchors however overlap with most ground truth boxes, leading to a huge imbalance between positive and negative anchor boxes and adds extra computational cost, which slows down training and inference~\cite{law2018cornernet}, \cite{dong2020centripetalnet}. Third, anchor boxes must be designed carefully in terms of their number, scales, and aspect ratios, as varying these parameters impacts performance. 

\begin{figure}[t]
\begin{center}
\includegraphics[width=.9\linewidth]{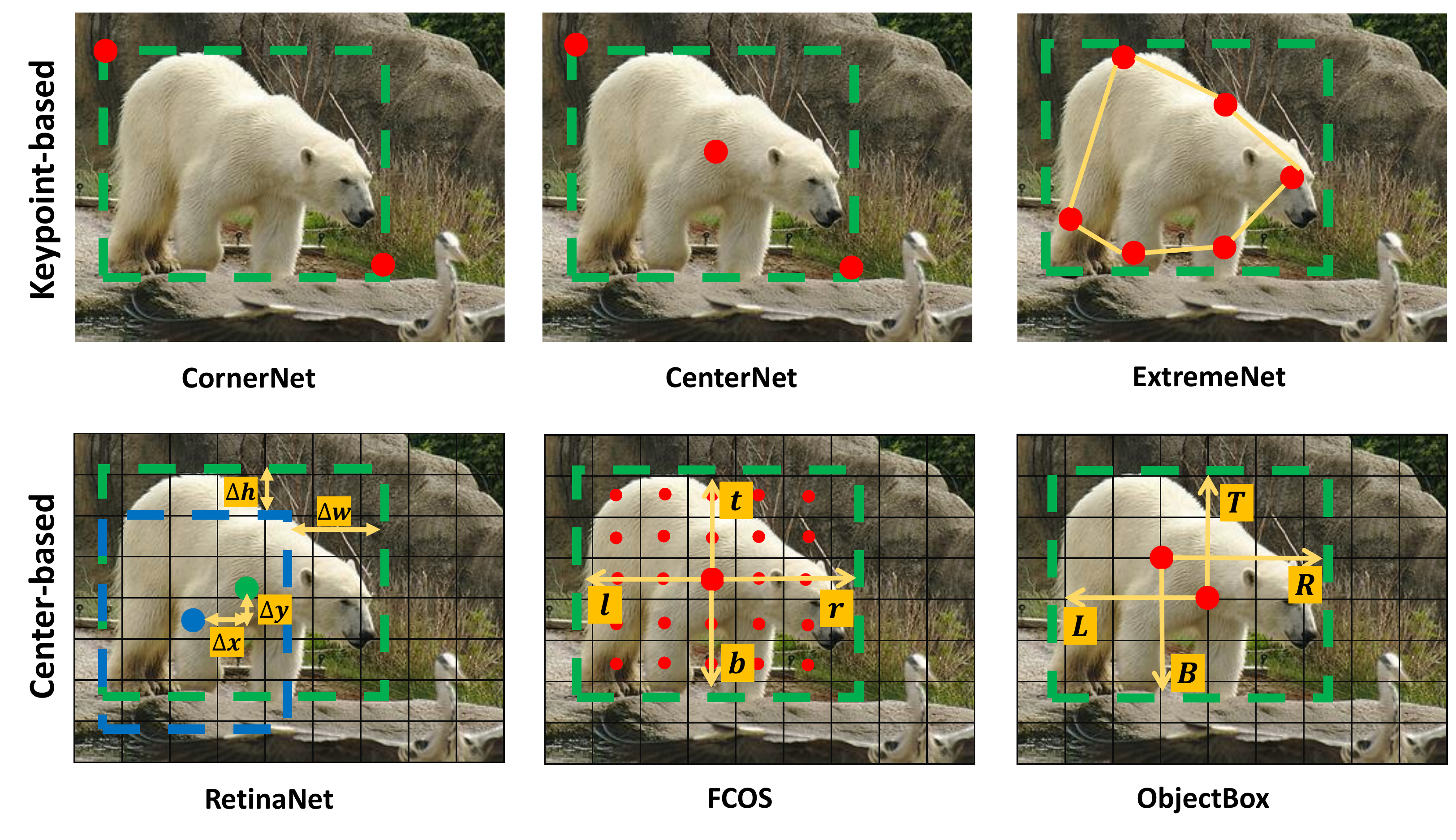}
\end{center}
   \caption{The first row shows keypoint-based anchor-free methods which use different combinations of keypoints and then group them for bounding box prediction. A pair of corners, a triplet of keypoints, and extreme points on the object are respectively used in CornerNet~\cite{law2018cornernet}, CenterNet~\cite{duan2019centernet}, and ExtremeNet~\cite{zhou2019bottom}. The second row shows center-based methods, which can be anchor-based (such as RetinaNet~\cite{lin2017focal}) or anchor-free (such as FCOS~\cite{tian2019fcos}). As opposed to FCOS which employs all the locations inside the bounding box, ObjectBox only uses 2 corners of the central cell location for bounding box regression }
   \label{fig:methods}
\end{figure}

In response to these challenges, a number of anchor-free object detectors~\cite{redmon2016you}, \cite{tian2019fcos}, \cite{law2018cornernet}, \cite{zhou2019bottom}, \cite{zhou2019objects}, \cite{huang2015densebox}, \cite{yu2016unitbox} have been recently developed, which can be categorized into keypoint-based~\cite{redmon2016you}, \cite{law2018cornernet}, \cite{zhou2019bottom}, \cite{zhou2019objects} and center-based methods~\cite{huang2015densebox}, \cite{tian2019fcos}, \cite{yu2016unitbox}. 
In keypoint-based methods, multiple object points, such as center and corner points, are located using a standard keypoint estimation network (e.g., HourglassNet~\cite{newell2016stacked}), and grouped to bound the spatial extent of objects. They however require a complicated combinatorial grouping algorithm after keypoint  detection. In contrast, center-based methods are more similar to anchor-based approaches as they use the object region of interest or central locations to define positive samples. 
While anchor-based methods use anchor boxes as predefined reference boxes on these central locations, anchor-free methods instead directly regress the bounding boxes at these locations (see Figure~\ref{fig:methods}). 

\begin{figure}[t]
\begin{center}
 \includegraphics[width=1\linewidth]{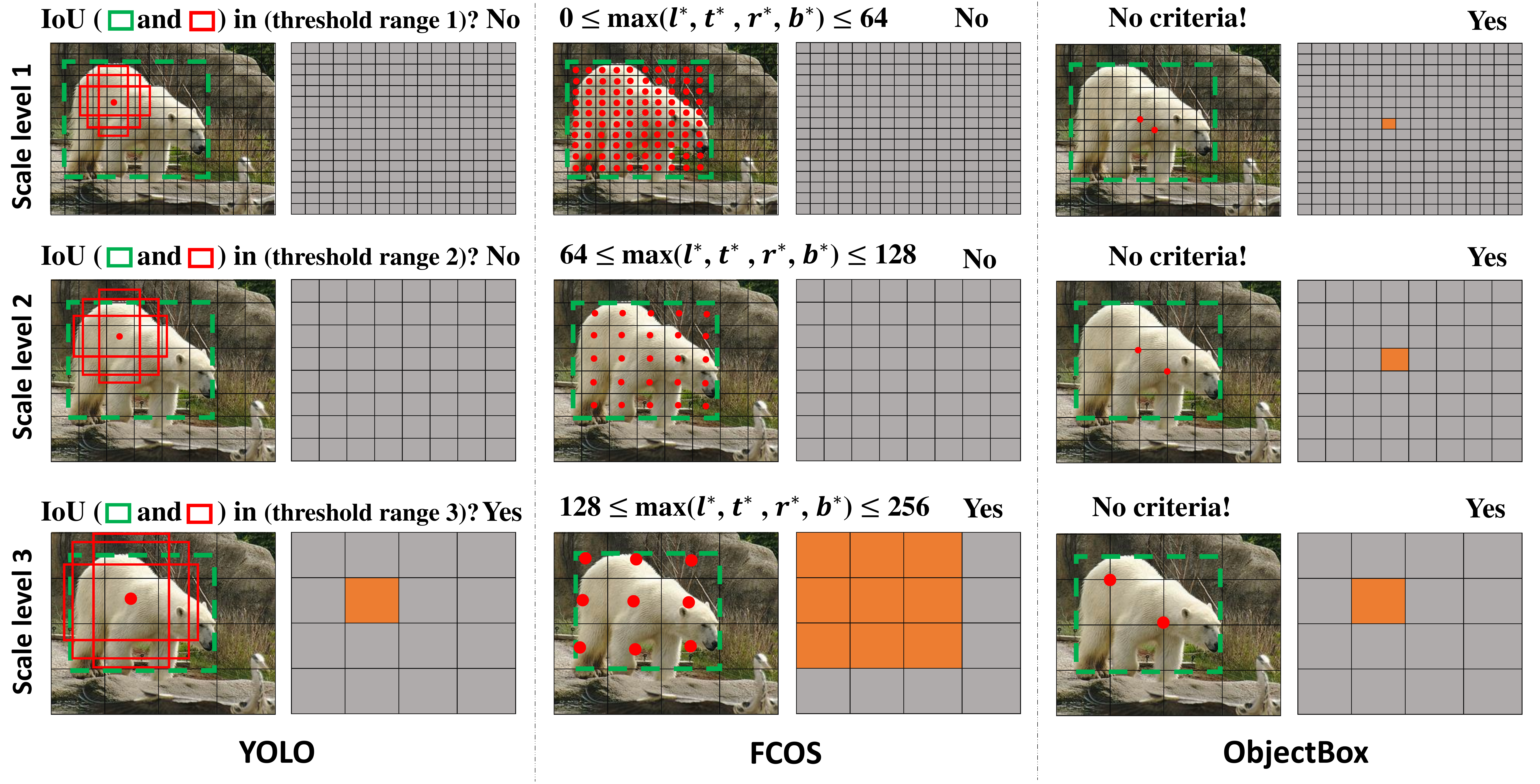}
\end{center}
   \caption{ObjectBox treats the target boxes at all scales as positive (orange) samples, while target boxes at some scales are discarded as negatives (gray) in other methods (both anchor-based and anchor-free). For instance, YOLO utilizes the IoU scores to threshold out negative samples and FCOS uses range constraints to select positive samples
   }
   \label{fig:scales}
\end{figure}

It is shown in~\cite{zhang2020bridging} that the main difference between anchor-based and anchor-free methods in center-based approaches is the definition of positive and negative training samples, which leads to a performance gap. To distinguish between positive and negative samples, anchor-based methods use IoU to select positives in spatial and scale dimension simultaneously, whereas anchor-free methods use some spatial and scale constraints to first find candidate positives in the spatial dimension, then select final positives in the scale dimension. 
Nevertheless, both static strategies impose constraint thresholds
to determine the boundaries between positive and negative samples, ignoring the fact that for objects with different sizes, shapes or occlusion conditions, the optimal boundaries may vary~\cite{ge2021ota}. 
Many dynamic assignment mechanisms have been developed in response to this issue~\cite{zhang2020bridging}, \cite{ge2021ota}, \cite{kim2020probabilistic}. For instance, in~\cite{zhang2020bridging}, the division boundary is proposed to be set for each target based on some statistical criterions.

In this paper, we propose to relax all constraints imposed by static or dynamic assignment strategies and, thus, treat all objects in all scales equally. To learn the classification labels and regression offsets regardless of the object shape or size, we only regress from object central locations which are treated as shape- and size- agnostic anchors~\cite{zhou2019objects}. To support this, we define new regression targets as the distances from two corners of the grid cell that contains the object center,  to  the bounding  box  boundaries ($L$, $R$, $B$, and $T$ in Figure~\ref{fig:methods}). 
As illustrated in Figure~\ref{fig:scales}, we use no criteria compared to other methods in different scale levels. We therefore expand the positive samples without any bells and whistles. To learn these positive samples from all scales, we propose a new scale-invariant criteria as an IoU measure which penalizes the error between target and predicted object boxes with different sizes at different scale levels.

In summary, our contribution is the proposal of a novel anchor-free object detector, ObjectBox, which is better equipped to handle the label assignment issue, and performs favorably in comparison to
the state-of-the-art. Moreover, our method is plug-and-play and can be easily applied across various datasets without the need for any hyperparameter tuning. Our method is therefore more robust and generalizable, and achieves state-of-the-art results. Lastly, we will make our code implementation publicly available upon publication of this paper.

\section{Related Work}

\subsection{Anchor-based object detectors}
To localize objects at different scales with various aspect ratios, Faster R-CNN introduced \emph{anchor boxes} as fixed sized bounding box proposals. 
The rationale behind anchor boxes is to use a set of predefined shapes (i.e. sizes and aspect ratios) as bounding box proposals, an idea which has become common in other object detection methods~\cite{redmon2018yolov3}, \cite{bochkovskiy2020yolov4}, \cite{lin2017focal}, \cite{liu2016ssd}. 

Early anchor-based methods include two stages for region proposal generation and object detection, which make them unsuitable for real-time applications. To achieve real-time performance, single-shot detectors~\cite{lin2017focal}, \cite{liu2016ssd}, \cite{redmon2018yolov3}, \cite{yi2019assd} used anchors without relying on RPNs. They directly predicted bounding boxes and class probabilities from the entire image in a single evaluation. The most representative single-shot detectors are SSD~\cite{liu2016ssd}, RetinaNet~\cite{lin2017focal}, and YOLO~\cite{redmon2018yolov3}, \cite{bochkovskiy2020yolov4}. 
Several other techniques used different variations of anchor boxes. For example, 
a multiple anchor learning approach was proposed in~\cite{ke2020multiple} to construct anchor bags and select the most representative anchors from each bag.

\subsection{Anchor-free object detectors}
A limitation of anchor-based methods is that they
require predefined hyperparameters to specify the sizes and
aspect ratios of the anchor boxes.
Specifying these hyperparameters requires heuristic tuning and several empirical tricks, and is dependent on the dataset and therefore lacks generality. Anchor-free detectors have been recently proposed to overcome the drawbacks of anchor boxes. They can be categorized as \emph{keypoint-based} and \emph{center-based} approaches. 

\textbf{Keypoint-based methods}
detect specific object points, such as center and corner points, and group them for bounding box prediction. Although they show improved performance over anchor-based methods, the grouping procedure is time-consuming, and they usually result in a low recall rate. Some representative examples include CornerNet~\cite{law2018cornernet}, ExtremeNet~\cite{zhou2019bottom}, CenterNet~\cite{zhou2019objects}, \cite{duan2019centernet}, and
CentripetalNet~\cite{dong2020centripetalnet}.

\textbf{Center-based methods} 
use an object region of interest or central locations to determine positive samples, which makes them more comparable to anchor-based approaches. 
FCOS~\cite{tian2019fcos}, for instance, considered all locations within the object bounding box to be candidate positives and found the final positives in each scale dimension. It computed the distances from these positive locations to the four sides of the bounding box. It however generated many low-quality predicted bounding boxes from locations far from the object center. To suppress these predictions, it used a \emph{centerness} score to down-weight the scores of low-quality bounding boxes. Moreover, it utilized a 5-level FPN (Feature Pyramid Network)~\cite{lin2017feature}
to detect objects with different sizes at different levels of feature maps. 
FoveaBox~\cite{kong2020foveabox}
predicted both the locations where the object center is likely to exist, and the bounding box for each positive location. 
FSAF (Feature Selective Anchor-Free)~\cite{zhu2019feature} attached an anchor-free branch to each level of the feature pyramid in RetinaNet~\cite{lin2017focal}.

\subsection{Label assignment}

It is shown in~\cite{zhang2020bridging} that
anchor-based and anchor-free methods achieve similar results if they use the same \emph{label assignment} strategy. In label assignment, each feature map point is labeled positive or negative based on the object ground-truth and the assignment strategy. 
Some anchor-free methods such as FCOS~\cite{tian2019fcos} utilize static constraints to define positives, while a proper constraint may vary based on the objects' sizes and shapes. 

Many other label assignment strategies have been recently proposed. ATSS~\cite{zhang2020bridging} (Adaptive Training Sample Selection), for example, proposed a dynamic strategy based on statistical features of the objects.
In~\cite{kim2020probabilistic}, the anchor assignment is modeled as a probabilistic procedure by calculating anchor scores from a detector model and maximizing the likelihood of these scores for a probability distribution. OTA~\cite{ge2021ota} (Optimal Transport Assignment) proposed to formulate label assignment as an optimal transport problem, which is a variant of linear programming in optimization theory. It characterized each ground-truth as a \emph{supplier} of a particular number of labels, and defines each anchor as a \emph{demander} that requires one unit label. If an anchor obtains a large enough number of positive labels from a given ground-truth, it is treated as one positive anchor for that ground-truth.

These strategies, however, do not maintain \emph{equality between different objects}, and they tend to assign more positive samples for larger objects. This can be alleviated by assigning the same number of positive samples and allowing learning to occur at all scales for every object regardless of its size.

\section{ObjectBox}

Let a training image $X\!\in\!\mathbb{R}^{W\!\times \!H\!\times\!3}$ contain $n$ objects with ground-truth $\{b_i, c_i\}_{i=1}^n$, where $b_i$ and $c_i$ respectively denote the bounding box and the object class label for the $i$\textsuperscript{th} object. Each bounding box $b=\{x,y,w,h\}$ is represented by its center $(x,y)$, width $w$ and height $h$. Our goal is to locate these boxes in an image and assign their class labels.

\begin{figure}[t] 
\begin{center}
 \includegraphics[width=0.9\linewidth]{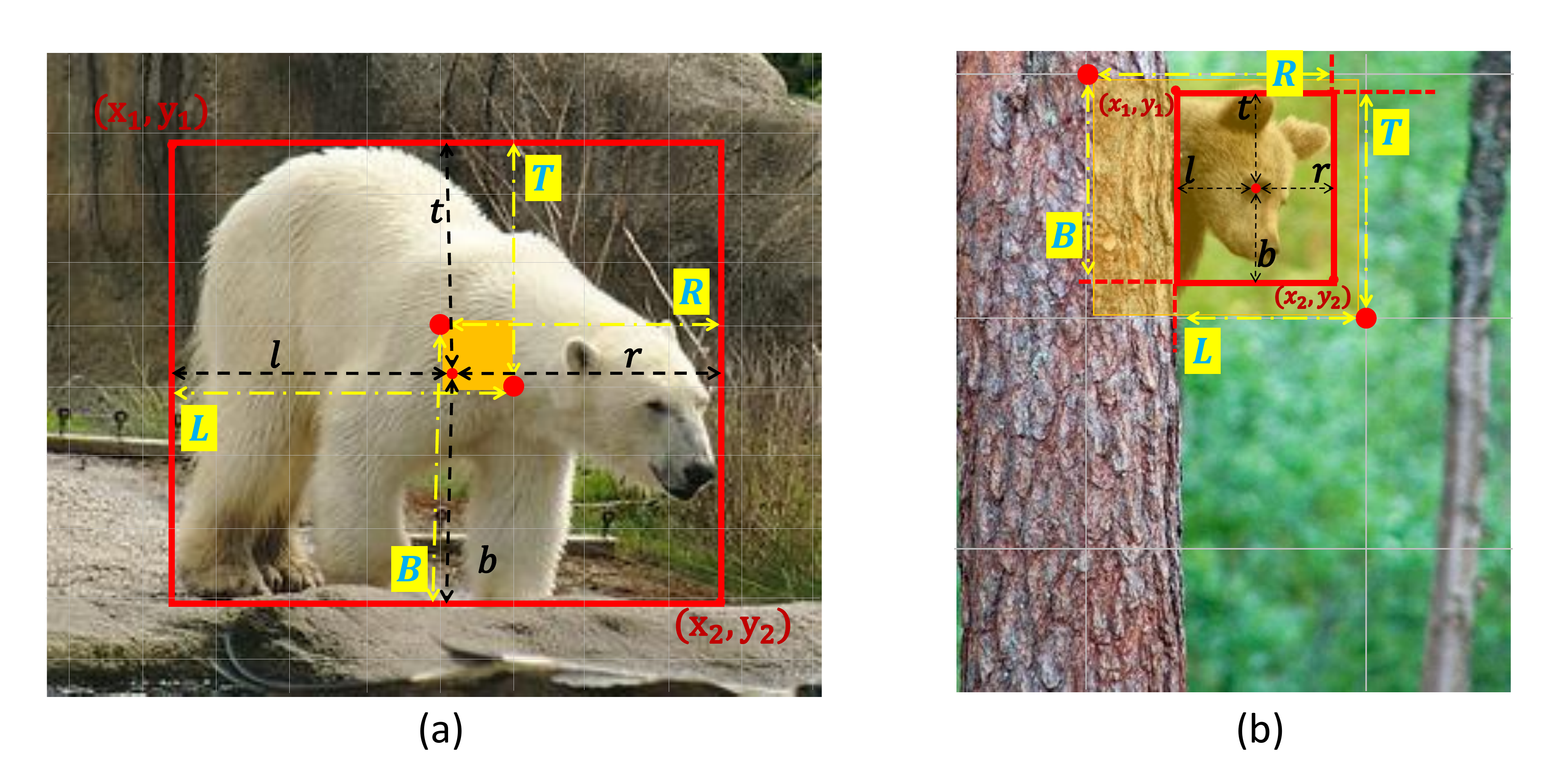}
\end{center}
   \caption{ObjectBox computes the distances from two corners of the center cell to the bounding box boundaries. A large and small object are respectively shown in (a) and (b). In (b), the small object lies completely within a cell, which usually occurs in larger strides (e.g., $s_i=32$). ObjectBox however does not discard these cases as it regresses to four sides of the bounding box for all objects with varying scales
   }
   \label{fig:bbox}
\end{figure}

\subsection{Label assignment based on object central locations}

The bounding box $b$ with center $(x,y)$ in the input image can be defined using its corner points as $\{(x_1^{(i)},y_1^{(i)}),(x_2^{(i)},y_2^{(i)})\}$, where $(x_1^{(i)},y_1^{(i)})$ and $(x_2^{(i)},y_2^{(i)})$ denote the 
respective coordinates of the top-left and bottom-right corners at scale $i$. Our method predicts bounding boxes at 3 different scales to handle object scale variations. Hence, different sizes of objects can be detected on 3 feature maps corresponding to these scales. We specifically choose strides $s\!=\!\{8,16,32\}$ and map each bounding box center to certain locations on these embeddings. 

We map the center $(x,y)$ to the center location 
(i.e., the orange cell in Figure~\ref{fig:bbox} (a)) in the embedding for scale $i$, and separately compute the distances from its top-left and bottom-right corners (red circles) each respectively from two boundaries of the bounding box. Specifically, as shown in Figure~\ref{fig:bbox}, we compute the distances from the bottom-right corner to the left and top boundaries ($L$ and $T$), and the distances from the top-left corner to the right and bottom boundaries ($R$ and $B$) as follows:
\begin{equation} 
\begin{cases}
 L^{{(i)}^*}=(\lfloor \frac{x}{s_i}\rfloor+1) - (x_1^{(i)}/s_i) \\
 T^{{(i)}^*}= (\lfloor \frac{y}{s_i}\rfloor+1) - (y_1^{(i)}/s_i) \\
 R^{{(i)}^*}= (x_2^{(i)}/s_i) - \lfloor \frac{x}{s_i}\rfloor \\ B^{{(i)}^*}= (y_2^{(i)}/s_i) - \lfloor \frac{y}{s_i}\rfloor \\
\end{cases}\label{eq:dist}
\end{equation}
where $(L^{{(i)}^*},T^{{(i)}^*},R^{{(i)}^*},B^{{(i)}^*})$ represent the regression targets at scale $i$, and $(\lfloor \frac{x}{s_i}\rfloor, \lfloor \frac{y}{s_i}\rfloor)$ and $(\lfloor \frac{x}{s_i}\rfloor + 1, \lfloor \frac{y}{s_i}\rfloor + 1)$ denote the respective coordinates of the top-left and the bottom-right corners of the center location. It should be noted that $L^{{(i)}^*}\!+\!R^{{(i)}^*}\!=\!w^{(i)}\!+\!1$ and $T^{{(i)}^*} + B^{{(i)}^*} = h^{(i)} + 1$, where $w^{(i)}=w/s_i$ and $h^{i}=h/s_i$ denote the width and height of the bounding box $b$ at scale $i$, respectively. The predictions corresponding to these distances are as follows:


\begin{equation}
\begin{cases}
    L^{(i)}=(2 \times \sigma (p_0))^2 * 2^i \\
    T^{(i)}= (2 \times \sigma (p_1))^2 * 2^i \\
    R^{(i)}= (2 \times \sigma (p_2))^2 * 2^i \\
    B^{(i)}= (2 \times \sigma (p_3))^2 * 2^i \\
\end{cases}
\label{eq:pred}
\end{equation}
where $\sigma$ stands for the logistic sigmoid function, and $(p_0, p_1, p_2, p_3)$ denote the network predictions for distance values, which we enforce by sigmoid, to be in the range of 0 and 1. Multiplying by 2 allows detected values to cover a slightly larger range. With $()^2$, the output is stably initialized with around zero gradient. We differentiate between different scales by multiplying to a constant scale gain, \emph{i.e.}, $2^i, i= 1, 2, 4$. The overall network outputs include one prediction per location per scale, each of which comprises the above-mentioned distance values, as well as an objectness score and a class label for each bounding box.


Our formulation ensures that all the distances being regressed remain positive under different conditions. As illustrated in Figure~\ref{fig:bbox} (b), the 4 distances can be computed as positive values even for a small object which is contained completely within a cell at a larger stride.
More importantly, we treat all the objects as positive samples at different scales. This is in contrast to existing center-based approaches (i.e., both anchor-based and anchor-free methods). In the anchor-based methods, for instance, each center location in a certain scale is seen as the center of multiple anchor boxes, and
if the IoUs of the target box and these anchor boxers are not within the threshold ranges, then it is considered as a negative sample. Similarly, anchor-free methods discard some target boxes as being negative samples based on different spatial and scale constraints. FCOS~\cite{tian2019fcos}, for example, defines a set of maximum distance values that limit the range of object sizes that can be detected at each feature level.
As another example, FoveaBox~\cite{kong2020foveabox} controls the scale range for each pyramid level by an empirically-learned parameter, while in~\cite{zhu2019feature}, a set of constant scale factors is used to define positive and negative boxes. As seen
in Figure~\ref{fig:scales}, ObjectBox however treats all target boxes at all scales as positive samples. It therefore learns from all scales regardless of the object size to achieve more reliable regressions from multiple levels. As ObjectBox considers only central locations for each object, the number of positive samples per object is independent of object size.

As the geometric center of the box might lie near a boundary of the center cell, we augment the center with its neighboring cells. For example, the above location is used in addition to the center cell when the center of the bounding box is on the upper half of the cell. 

Our method detects the objects from their central regions. If two boxes overlap, their centers are less likely to overlap given that it is quite rare for two box centers to be situated at the same location. In MS-COCO~\cite{lin2014microsoft} and PASCAL VOC 2012~\cite{everingham2015pascal}, we found no cases where centers of overlapping objects overlap. Our augmented center locations, however, can be useful in dealing with these boxes. 
In our experiments (Section~\ref{sec:Ablation}), we show that adding more points in addition to the central locations hurts the detection performance.

Our strategy implicitly harnesses the intuition behind anchor boxes, which are usually created by clustering the dimensions of the ground truth boxes in the dataset~\cite{redmon2017yolo9000}. 
Their dimensions are obtained as estimates of the most common shapes in different sizes. For instance, Faster R-CNN~\cite{ren2015faster} and YOLO~\cite{redmon2018yolov3,bochkovskiy2020yolov4} use three scales and three aspect ratios, yielding 9 anchors at each position. Our method however uses the central locations of the bounding boxes at each scale 
to generate multiple predictions for each object. Our method is also more effective than other anchor-free methods such as FCOS~\cite{tian2019fcos} which leverage additional levels of FPN (i.e., a total of 5 layers) to handle the overlapping bounding boxes.

\begin{figure}[t] 
\begin{center}
 \includegraphics[width=0.8\linewidth]{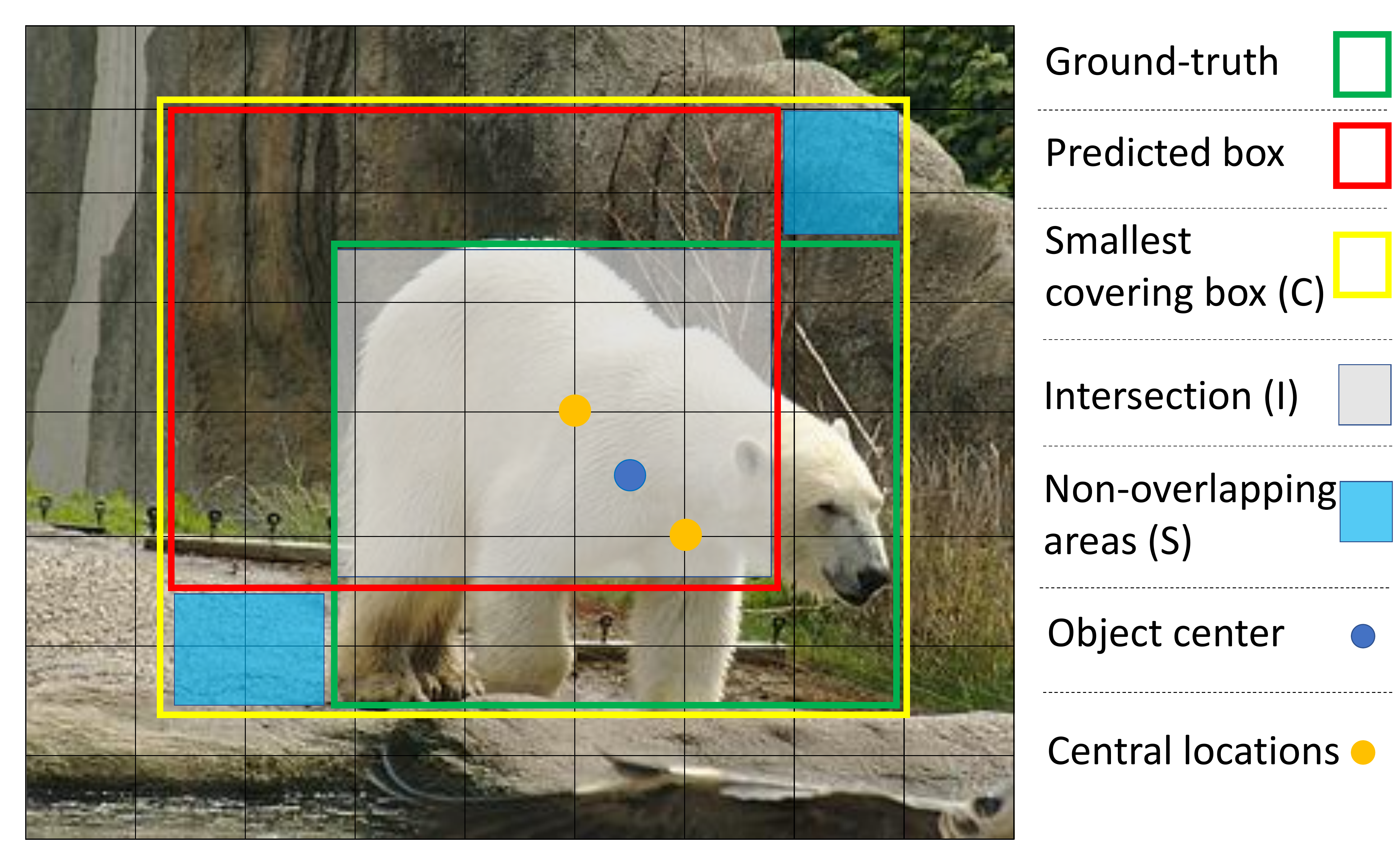}
\end{center}
   \caption{The areas in SDIoU loss for box regression}
   \label{fig:our_loss}
\end{figure}

\subsection{Box Regression}

As $\{L^{(i)}, T^{(i)}, R^{(i)}, B^{(i)}\}$ are distances, they can be treated independently and Mean Square Error (MSE) can be used to perform regression on these values individually. Nevertheless, such a strategy would disregard the integrity of the object bounding box. IoU (Intersection over Union or Jaccard index) loss has already been proposed to take the coverage of the predicted and ground-truth bounding box areas into consideration. IoU is a widely-used similarity metric between two shapes, which due to its appealing feature of being differentiable, can be directly used as an objective function for optimization~\cite{yu2016unitbox,rezatofighi2019generalized,zheng2020distance,tychsen2018improving}. In object detection, IoU can encode the width, height, and location of each bounding box into a normalized measure.
The IoU loss ($\mathcal{L}_{IoU} = 1 - IoU$) thus allows
a bounding box
to be recognized  as a single entity, and jointly regresses the four coordinate points of the bounding box. 

IoU loss has been recently improved
upon by considering different cases. For example, GIoU (Generalized IoU) loss~\cite{rezatofighi2019generalized} included the shape and orientation of the object in addition to the coverage area. It can find the smallest area that can simultaneously cover the predicted and ground-truth bounding boxes, and use it as the denominator to replace the original denominator used in IoU loss. DIoU (Distance IoU) loss~\cite{zheng2020distance} additionally emphasized the distance between the centers of the predicted and ground-truth boxes. CIoU (Complete IoU) loss~\cite{zheng2020distance} simultaneously included the overlapping area, the distance between center points, and the aspect ratio.

In our case, we are interested in minimizing the distance between two boxes which are each given by four distance values. As we learn from different scales for objects with different sizes (\emph{i.e.}, we do not differentiate between scale levels), our bounding box regression loss function should be scale-invariant. Nevertheless, $\ell_n$-based losses grow as the scales of the bounding boxes become larger~\cite{sun2020scale}.
As opposed to the original IoU loss and its variants, our loss does not require the bounding box locations to be matched, since the localization task is already embedded in the process.
Moreover, the predicted and ground-truth boxes share at least one point in the worst case (i.e., overlap $\ge 0$). This is because $\{L^{(i)},\!T^{(i)},\!R^{(i)},\!B^{(i)} \} \!\ge \!0$ for each box. In this work, we propose an IoU-based loss tailored for our object detection method, which can be used to improve other anchor-free detectors as well (the experiments are provided in the supplementary materials). Our proposed loss, called SDIoU which stands for scale-invariant distance-based IoU, is directly applied on the network outputs which are distance values from the object center to top-left and bottom-right corners. Other IoU-based losses, however, work on the object center and object width and height. As SDIoU is based on the Euclidean distances between corresponding offsets of the predicted and ground-truth boxes, it can keep the box integrity and score the overlapping area in all 4 directions.

Similar to CIoU~\cite{zheng2020distance} and scale balanced loss~\cite{sun2020scale}, we consider non-overlapping areas, overlapping or intersection area, and smallest box that covers both boxes. 
We first compute the non-overlapping area, $S$, by summing the squares of all the Euclidean distances between corresponding distance values as:
\begin{equation}
    S = (L^* - L)^2 + (T^* - T)^2 + (R^* - R)^2 + (B^* - B)^2 ,
\end{equation}
where $\{L,T,R,B\}$ and $\{L^*,T^*,R^*,B^*\}$ are the predicted and ground-truth distances, respectively. (We omit here the scale $i$, for better readability.) Intuitively, computing the squared Euclidean distances between different distance values can effectively consider the predicted and ground-truth distances at 4 directions.

We obtain the intersection area, $I$, by computing the square of the length of the intersection area's diagonal as:
\begin{equation}
    I = (w^I)^2 + (h^I)^2,
\end{equation}
where $w^I$ and $h^I$ are the width and height of the intersection area, respectively, and are computed as:
\begin{equation}
\begin{aligned}
    &w^I = min(L^*, L) + min(R^*, R) - 1 \\
    &h^I = min(T^*, T) + min(B^*, B) - 1. \\
\end{aligned}
\end{equation}

The smallest area that covers both predicted and ground-truth boxes, $C$, is calculated by the square of its length as:
\begin{equation}
    C = (w^C)^2 + (h^C)^2,
\end{equation}
where $w^C$ and $h^C$ respectively denote $C$'s width and height, which are computed as:
\begin{equation}
\begin{aligned}
    &w^C = max(L^*, L) + max(R^*, R) - 1 \\
    &h^C = max(T^*, T) + max(B^*, B) - 1. \\
\end{aligned}
\end{equation}    
By minimizing $C$, the predicted box can move towards the ground-truth box at 4 directions. We finally compute the SDIoU as:
\begin{equation}
    SDIoU = \frac{(I-\rho S)}{C},
\end{equation}
where $\rho$ denotes a positive trade-off value that favors the overlap area (we however set $\rho = 1$ in all the experiments). We use both $I$ and $(-S)$ in the numerator to score the intersection area as well as penalizing the non-overlapping area. The predicted 4 distance values are thus enforced to faster match the ground-truth distances. The SDIoU loss is eventually defined as $\mathcal{L}_{IoU} = 1-IoU$. Figure~\ref{fig:our_loss} illustrates the areas considered in our SDIoU loss.

\section{Experiments}
\label{sec:Experiments}
\noindent \textbf{Datasets.} Two common challenging datasets, MS-COCO~\cite{lin2014microsoft} and PASCAL VOC 2012~\cite{everingham2015pascal}, which are widely-used benchmarks for natural scene object detection, were selected to evaluate the proposed ObjectBox method and compare it against current state-of-the-art methods. MS-COCO is a challenging dataset that includes a large number of objects labeled in 80 object categories. We used the trainval35k split containing 115k images for training  our network, and reported the results on the test-dev split with 20k images.
The PASCAL VOC 2012 dataset consists of complex scene images of 20 diverse object classes. We trained our model using the VOC 2012 and VOC 2007 trainval splits (17k images) and tested it on the VOC 2012 test split (16k images). Experimental results on the PASCAL VOC 2012~\cite{everingham2015pascal} can be found in the supplementary materials (Sec. S.2).

\noindent \textbf{Implementation Details.} We implemented our method on two different backbones, \emph{i.e.}, ResNet-101 and  CSPDarknet~\cite{wang2020cspnet}, \cite{huang2017densely}, \cite{bochkovskiy2020yolov4}. We use ResNet-101 which is a widely-used backbone in many object detectors to provide a fair comparison with other state-of-the-art methods. 
We also utilized CSPDarknet and add SPP (Spatial Pyramid Pooling)~\cite{he2015spatial}, \cite{redmon2018yolov3}, \cite{bochkovskiy2020yolov4} over the backbone to increase the receptive field of the extracted features. CSPDarknet has the potential to enhance the learning abilities of the CNNs and reduce the memory cost~\cite{bochkovskiy2020yolov4}. 

The training hyperparameters were set to an initial learning rate of 0.01, momentum of 0.937, weight decay of 0.0005, warm-up epochs of 3, and warm-up momentum of 0.8.  The initial learning rate was multiplied with a factor 0.1 at 400,000 steps, and then again at 450,000 steps. We set the batch size to 24 and used SGD optimization. We trained our models to a maximum of 300 epochs with early stopping patience of 30 epochs. The experiments were executed on a single Titan RTX GPU. The NMS (Non-Maximum Suppression) threshold was also set as $0.6$ in all experiments. 

We used CutMix and Mosaic data augmentation during training~\cite{bochkovskiy2020yolov4}. They both mix different contexts to facilitate detection of objects outside their normal context. CutMix mixes 2 input images, while Mosaic mixes 4 training images. 
For each scale level $s$, we use a multitask loss as:
\begin{equation}
    \ell^s =  \ell_{cls}^s + \ell_{obj}^s + \ell_{box}^s
\end{equation}
where $\ell_{cls}^s$, $\ell_{obj}^s$, and $\ell_{box}^s$ respectively denote the classification loss, 
a binary cross entropy loss,
and the regression loss for box offsets at scale $s$. We use the binary cross entropy between the target classes and the predicted probabilities as our classification loss and the binary confidence score. We employ SDIoU loss as the regression loss between the proposed targets and the predicted ones. The losses are computed for each scale and are summed 
as $\mathcal{L}=\sum_s \ell^s$.

\begin{table*}[t] 
\caption{Performance comparison with the state-of-the-art methods on the MS-COCO dataset in single-model and single-scale results. The bold and underlined numbers respectively indicate the best and second best results in each column}
    \resizebox{\textwidth}{!}{
    \centering
    \setlength
    \tabcolsep{1 pt}
    \begin{tabular}{llccccccccccccc}
        \hline
        & & \multicolumn{3}{c}{\underline{Avg. Precision, IoU}} & \multicolumn{3}{c}{\underline{Avg. Precision, Area}} & \multicolumn{3}{c}{\underline{Avg. Recall, \# Dets}} & \multicolumn{3}{c}{\underline{Avg. Recall, Area}} \\ 
        Method & Backbone & 
        AP & $AP_{50}$ & $AP_{75}$ & $AP_S$ & $AP_M$ & $AP_L$ & $AR_1$ & $AR_{10}$ & $AR_{100}$ & $AR_{S}$ & $AR_{M}$ & $AR_{L}$ \\
        \hline \hline
        
        SSD513~\cite{liu2016ssd} & ResNet-101 &  
        31.2 & 50.4 & 33.3 & 10.2 & 34.5 & 49.8 & 28.3 & 42.1 & 44.4 & 17.6 & 49.2 & 65.8 \\
        
        DeNet~\cite{tychsen2017denet} & ResNet-101 &
        33.8 & 53.4 & 36.1 & 12.3 & 36.1 & 50.8 & 29.6 & 42.6 & 43.5 & 19.2 & 46.9 & 64.3 \\
        
        F-RCNN w/ FPN~\cite{lin2017feature} & ResNet-101 &  
        36.2 & 59.1 & 39.0 & 18.2 & 39.0 & 48.2 & - & - & - & - & - & - \\
        
        YOLOv2~\cite{redmon2017yolo9000} & DarkNet-19 & 
        21.6 & 44.0 & 19.2 & 5.0 & 22.4 & 35.5 & 20.7 & 31.6 & 33.3 & 9.8 & 36.5 & 54.4 \\
        
        RetinaNet~\cite{lin2017focal} & ResNet-101 & 
        39.1 & 59.1 & 42.3 & 21.8 & 42.7 & 50.2 & - & - & - & - & - & - \\
        
        YOLOv3~\cite{redmon2018yolov3} & DarkNet-53 & 
        33.0 & 57.9 & 34.4 & 18.3 & 35.4 & 41.9 & - & - & - & - & - & - \\
        
        CornerNet~\cite{law2018cornernet} & Hourglass-104 &
        40.6 & 56.4 & 43.2 & 19.1 & 42.8 & 54.3 & \underline{35.3} & 54.7 & 59.4 & 37.4 & 62.4 & \textbf{77.2}  \\
        
        CenterNet~\cite{duan2019centernet} & Hourglass-52 &
        41.6 & 59.4 & 44.2 & 22.5 & 43.1 & 54.1 & 34.8 & {55.7} & {60.1} & {38.6} & {63.3} & {76.9} \\
        
        ExtremeNet~\cite{zhou2019bottom} & Hourglass-104 & 
        40.2 & 55.5 & 43.2 & 20.4 & 43.2 & 53.1 & - & - & - & - & - & -   \\
        
        FCOS~\cite{tian2019fcos} & ResNeXt-101 & 
        42.1 & 62.1 & 45.2 & 25.6 & 44.9 & 52.0 & - & - & - & - & - & - \\
        
        ASSD513~\cite{yi2019assd} & ResNet101 & 
        34.5 & 55.5 & 36.6 & 15.4 & 39.2 & 51.0 & 29.9 & 45.6 & 47.6 & 22.8 & 52.2 & 67.9 \\
        
        SaccadeNet~\cite{lan2020saccadenet} & DLA-34-DCN &
        40.4 & 57.6 & 43.5 & 20.4 & 43.8 & 52.8 & - & - & - & - & - & - \\
        
        YOLOv4~\cite{bochkovskiy2020yolov4} & CSPDarknet &
        43.5 & \underline{65.7} & 47.3 & {26.7} & 46.7 & 53.3 & - & - & - & - & - & - \\
        
        FoveaBox~\cite{kong2020foveabox} & ResNeXt-101 &
        {43.9} & 63.5 & {47.7} & \underline{26.8} & {46.9} & {55.6} & - & - & - & - & - & - \\
        
        RetinaNet+CBAF~\cite{tang2021coordinate} & ResNet-101 & 
        43.0 & 63.2 & 46.3 & 25.9 & 45.6 & 51.4 & - & - & - & - & - & - \\
        
        ATSS
        ~\cite{zhang2020bridging} & ResNet-101 & 
        43.6 & 62.1 & 47.4 & 26.1 & 47.0 & 53.6 & - & - & - & - & - & - \\
        
        PAA
        ~\cite{kim2020probabilistic} & ResNet-101 & 
        44.8 & 63.3 & 48.7 & 26.5 & 48.8 & 56.3 & - & - & - & - & - & - \\
        
        OTA
        ~\cite{ge2021ota} & ResNet-101 & 
        45.3 & 63.5 & 49.3 & 26.9 & 48.8 & 56.1 & - & - & - & - & - & - \\
        
        VarifocalNet~\cite{zhang2021varifocalnet} & ResNet-101 & 
        46.0 & 64.2 & \textbf{50.0} & \textbf{27.5} & \underline{49.4} & 56.9 & - & - & - & - & - & - \\
        
        ObjectBox & ResNet-101 & 
        \underline{46.1} & {65.0} & 48.3 & {26.0} & {48.7} & \underline{57.3} & \underline{35.3} & \underline{57.1} & \underline{60.5} & \underline{39.2} & \underline{65.0} & {76.9} \\
        
        ObjectBox & CSPDarknet & 
        \textbf{46.8} & \textbf{65.9} & \underline{49.5} & \underline{26.8} & \textbf{49.5} & \textbf{57.6} & \textbf{36.0} & \textbf{57.5} & \textbf{60.7} & \textbf{39.4} & \textbf{65.2} & \underline{77.0} \\

        \hline
    \end{tabular}}
    \label{tab:coco_results}
\end{table*}

\subsection{MS-COCO Object Detection}
Table~\ref{tab:coco_results} shows the 
evaluation results on the MS-COCO dataset. Compared to the baseline methods, ObjectBox is considerably more accurate, achieving the best AP performance of $46.8\%$ with a CSPDarknet backbone. Our method also achieves the second-best performance of $46.1 \%$ with a ResNet-101 backbone. The relative improvement of $AP$ (which is averaged over 10 IoU thresholds of $0.5$ to $0.95$) indicates that ObjectBox generates more accurate boxes with better localization. With the CSPDarknet backbone, the improvements are also achieved over 8 other metrics including $AP_{50}$, $AP_{M}$, $AP_{L}$, $AR_{1}$, $AR_{10}$, $AR_{100}$, $AR_{S}$, and $AR_{M}$. Notably, ObjectBox with ResNet-101 obtains the second-best performance over 7 different metrics. 
These improvements over both anchor-based and anchor-free methods are mainly due to our strategy to learn object features in different scales fairly. Nonetheless, this is  not possible without regressing from the object central locations, which can be seen as shape- and size-agnostic anchors. 

The relative improvement in $AR_S$ indicates that our method can detect more small objects (which are more likely to overlap and generally harder to detect). The performance boost is also evident for $AP_L$ when detection of larger objects can benefit from all feature maps at 3 scale levels. This is another major difference with other detectors which learn from all points in the objects. To maintain the relative equality between different objects,  they consider the larger objects as positive samples only for embeddings with larger strides. 

The second best performing method, VarifocalNet~\cite{zhang2021varifocalnet}, replaces the classification score of the ground-truth class with a new IoU-aware classification score. It is built on an ATSS~\cite{zhang2020bridging} version of FCOS~\cite{tian2019fcos}. In ATSS, the Adaptive Training Sample Selection (ATSS) mechanism is used to define positive and negative points on the feature pyramids during training. 
FoveaBox~\cite{kong2020foveabox}, which is also an anchor-free detector and concentrates on the object center, achieves $AP=43.9$. It however separates samples as positives and negatives at each scale.  
Improvements over FCOS~\cite{tian2019fcos} ($+4 \%$) shows that central regions of the objects include enough recognizable visual patterns to detect the objects if we consider positive samples from all scales, and therefore, learning all the pixels inside the bounding box is not required for a general object detection method. 

It is also interesting to note that ObjectBox does not use any data-dependent hyperparameters. Other anchor-free methods which tend to address the generalization issue often use a number of such hyperparameters. FCOS~\cite{tian2019fcos}, for example, defines a hyperparameter for thresholding the object sizes at different scales, while FoveaBox~\cite{kong2020foveabox} defines a hyperparameter to control the scale range.

\begin{table*}[!t] 
\caption{The ablation study of ObjectBox with CSPDarknet on MS-COCO. We investigate the influence of box regression from different locations (A), number of predictions per location per scale (B), and imposing constraints based on the object size (C) }
    \centering
    \begin{tabular}{l|l|lcccccc}
        \hline
         & &  & \multicolumn{3}{c}{\underline{Avg. Precision, IoU}} & \multicolumn{3}{c}{\underline{Avg. Precision, Area}}  \\ 

        & Experiment & Method & AP & $AP_{50}$ & $AP_{75}$ & $AP_S$ & $AP_M$ & $AP_L$ \\
        
        \hline \hline
        
        \multirow{6}{*}{A} & \multirow{6}{*}{\shortstack{Regression \\ locations}} & (1) center & 33.1 & 56.8 & 36.0 & 17.5 & 35.2 & 42.1 \\
        
        & & (2) aug. center (ObjectBox) & 46.8 & 65.9 & 49.5 & 26.8 & 49.5 & 57.6 \\
        
        & & (3) h-centers  & 42.3 & 56.9 & 46.5 & 24.1 & 45.3 & 54.2  \\
        
        & & (4) aug. center + h-centers & 41.7 & 58.2 & 45.2 & 23.6 & 43.3 & 54.5 \\
        
        & & (5) 4 corners & 28.2 & 51.5 & 35.6 & 16.0 & 33.9 & 41.3 \\
        
        & & (6) 4 corners + center & 37.4 & 57.8 & 43.0 & 20.4 & 39.7 & 45.5 \\
        
        \hline
        
        \multirow{2}{*}{B} & \multirow{2}{*}{\#Pred.} & 1 prediction (ObjectBox) & 46.8 & 65.9 & 49.5 & 26.8 & 49.5 & 57.6 \\
        
        & & 4 predictions & 37.3 & 58.3 & 41.9 & 19.5 & 41.6 & 48.0 \\
        
        \hline
        
        \multirow{4}{*}{C} & \multirow{4}{*}{\shortstack{Scale \\ constraints}} & $m=\{0,32,64,\infty\}$ & 29.6 & 45.8 & 30.4 & 17.0 & 31.8 & 40.6 \\
        
         & & $m=\{0,64,128,\infty\}$ & 35.8 & 58.0 & 36.8 & 19.2 & 39.1 & 46.5 \\
        
        & & $m=\{0,128,256,\infty\}$ & 30.4 & 49.2 & 32.0 & 16.8 & 33.5 & 43.5 \\
        
        & & $m=\{0,256,512,\infty\}$ & 27.3 & 43.5 & 29.6 & 14.7 & 30.4 & 38.1 \\
        
        \hline
    \end{tabular}
    \label{tab:ablation}
\end{table*}

\subsection{Ablation Study}
\label{sec:Ablation}
To verify the effectiveness of our method, we performed several experiments with different settings on the MS-COCO dataset. We utilized ObjectBox with a CSPDarknet backbone in all ablation experiments. 

\noindent\textbf{Box regression locations.}
Table~\ref{tab:ablation} part A shows the impact of regression from different locations by choosing the boxes to be regressed from different locations. We defined 6 cases: (1) only one location at the center (referred to as `center'), (2) center location augmented with its neighboring locations (as done in ObjectBox, denoted by `aug. center'), (3) the centers of the connecting lines between the box center and two top-left and bottom-right box corner points (referred to as `h-centers'), (4) central locations in (2) plus all locations in (3) (denoted by `aug. center + h-centers'), (5) four corners of the bounding box, and (6) corner points in (5) plus the center location. 
The results show that using only the center cell is not sufficient for box regression. Another important point is that (3) outperforms (1), meaning that selection of two other points close to the center is better than only center point. Removing these two locations and considering only central locations in (2) even brings further improvements. Interestingly, in (4), no improvement is seen over (3). This indicates  not only that considering locations other than the central locations does not add valuable information, but also that doing so can actually degrade detection performance. The worst case occurs when we use only the corner points of the bounding box. 
While the performance is improved by the addition of one center location to the points in (5), the results are still far from those in (2), (3), and (4), where box regression is obtained only from points that are closer to the center locations.

\noindent\textbf{Number of predicted boxes.}
We analyzed the influence of the number of predictions per location, and reported the results in Table~\ref{tab:ablation} part B. In this experiment, we assigned 4 predictions to each location based on the offset of the object center in that location. Specifically, each location was divided into four equal finer locations, with one prediction given to each of them. When we predict 4 boxes at each location, surprisingly the performance degrades, confirming that our strategy of returning just one prediction per scale level is indeed beneficial.  

\noindent\textbf{Specialized feature maps.}
To show the impact of imposing constraints on the feature maps at different scales, we chose four sets of thresholds: (1) $m=\{0, 32, 64, \infty\}$, (2) $m=\{0, 64, 128, \infty\}$, (3) $m=\{0, 128, 256, \infty\}$, and (4) $m=\{0, 256, 512, \infty\}$. An object at scale $i$ is considered as a negative sample if $\{w, h\} < m_{i-1}$ or $\{w, h\} > m_{i}$ for $i=1,2,3$. The negative boxes thus are not regressed. This is similar to both anchor-based and anchor-free detectors. Specifically, anchor-free methods like YOLO~\cite{redmon2018yolov3}, \cite{bochkovskiy2020yolov4} assign anchor
boxes with different sizes to different feature levels, and anchor-free methods such as FCOS~\cite{tian2019fcos} directly limit the range of box regression for each level. The results in Table~\ref{tab:ablation} part C show the high sensitivity of the performance to these thresholds. Moreover, this experiment verifies our choice of considering embeddings in all scale levels for all objects, as thresholding the feature maps drastically hurts the results.

\noindent\textbf{Loss functions.}
To show the effectiveness of our SDIoU loss for box regression, we replaced it with three other common losses in three different experiments. We first used MSE (Mean Square Error) loss on all 4 distances separately. In the second and third experiments, we converted the 4 distances to $\{x,y,w,h\}$ and used the GIoU~\cite{rezatofighi2019generalized} and CIoU losses~\cite{zheng2020distance}. As observed in Table~\ref{tab:ablation_loss}, these losses are not suitable in anchor-free detectors like ObjectBox. More importantly, the benefit of our IoU loss is evident from these experiments. 

We provide more experiments in the supplemental materials (Sec. S.4) to verify the effectiveness of SDIoU in other anchor-free approaches like FCOS~\cite{tian2019fcos}.

\begin{table}[t] 
\caption{ The influence of different loss functions on ObjectBox}
    \centering
    \begin{tabular}{lcccccc}
        \hline
         & \multicolumn{3}{c}{\underline{Avg. Precision, IoU}} & \multicolumn{3}{c}{\underline{Avg. Precision, Area}}  \\ 

        Method & AP & $AP_{50}$ & $AP_{75}$ & $AP_S$ & $AP_M$ & $AP_L$ \\
        
        \hline \hline
        
        MSE & 22.6 & 44.1 & 19.4 & 12.5 & 18.3 & 35.7 \\
        
        Adopted GIoU & 27.4 & 46.9 & 28.2 & 23.8 & 30.2 & 41.8 \\
        
        Adopted CIoU & 27.1 & 46.5 & 28.1 & 24.0 & 30.5 & 41.0 \\
        
        SDIoU & 46.8 & 65.9 & 49.5 & 26.8 & 49.5 & 57.6 \\
        
        \hline
    \end{tabular}
    \label{tab:ablation_loss}
\end{table}

\section{Conclusion}
ObjectBox, an anchor-free object detector, is presented without the need for any 
hyperparameter tuning. 
It uses object central locations and employs a new regression target for bounding box regression. Moreover, by relaxing the label assignment constraints, it treats all objects equally in all feature levels. A tailored IoU loss also minimizes the distance between the new regression targets and the predicted ones. It was demonstrated that using existing backbone architectures such as CSPDarknet and ResNet-101, ObjectBox compares favorably to other anchor-based and anchor-free methods.

\paragraph
\noindent \textbf{Acknowledgments.} Thanks to Geotab Inc., the City of
Kingston, and the
Natural Sciences and Engineering Research Council of Canada (NSERC) for their support of this work.

\newpage

\bibliographystyle{splncs04}
\bibliography{egbib}
\end{document}


\pagestyle{headings}
\mainmatter
\def\ECCVSubNumber{6191}  

\title{ObjectBox: From Centers to Boxes for Anchor-Free Object Detection \\
(Supplementary Material)} 

\titlerunning{ObjectBox: From Centers to Boxes for Anchor-Free Object Detection}
%
\author{Mohsen Zand\orcidlink{0000-0001-8177-6000
} \and
Ali Etemad\orcidlink{0000-0001-7128-0220} \and
Michael Greenspan\orcidlink{0000-0001-6054-8770}}
\authorrunning{M. Zand et al.}
%
\institute{Dept. of Electrical and Computer Engineering, 
Ingenuity Labs Research Institute \\ Queen's University, Kingston, Ontario, Canada }
\maketitle

\section{Overview}
\label{sec:Overview}
We provide additional experiments to further explore the robustness of our method. 
Experimental results on the PASCAL VOC 2012~\cite{everingham2015pascal} are represented in Sec.~\ref{sec:pascal}. In Sec.~\ref{sec:multiscale}, we investigate correlations between the object size and the scale at which it is detected. 
We also utilize our IoU loss in other detectors and report the results in Sec.~\ref{sec:iou_loss}. Inference details are also discussed in Sec.~\ref{sec:inference}. Finally, we qualitatively show some ObjectBox results in Sec.~\ref{sec:visualization}.

\section{PASCAL VOC 2012} 
\label{sec:pascal}
To demonstrate the effectiveness of our method on different object categories in a subtle way, we perform another experiment on the PASCAL VOC 2012 dataset. 
We trained the network under the same settings as we performed on MS-COCO dataset. Notably, our method does not need to set dataset-dependent hyperparameters like anchor boxes. 
The results are shown in Table~\ref{tab:pascal_results}.  ObjectBox outperforms the other methods, achieving a higher AP score on 13 of the 20 object classes. 
Overall, ObjectBox achieves an mAP of 83.7$\%$, which is $+2.4 \%$ higher than the next best performing method. It can be observed that ObjectBox works relatively well in both small object and large object classes. For example, it achieves 92.1$\%$ in the class \emph{`plane'} and 93.3$\%$ in the class \emph{`car'}. It can be observed that ObjectBox is either the best or the second-best method in terms of AP score in all categories except ‘cat’, ’dog’, and ‘bike’. 
This is probably due
to the use of YOLO’s Darknet backbone, as YOLOv2 similarly does not work well in these categories.

\begin{table*}[t] 
\caption{Detection results on the PASCAL VOC 2012 dataset. F-RCNN denotes Faster R-CNN.  The  bold  and  underlined  numbers respectively indicate the best and second best results in each column}
    \resizebox{\textwidth}{!}{
    \setlength
    \tabcolsep{2pt}
    
    \begin{tabular}{lc|@{}c|@{}c|@{}c|@{}c|@{}c|@{}c|@{}c|@{}c|@{}c|@{}c|@{}c|@{}c|@{}c|@{}c|@{}c|@{}c|@{}c|@{}c|@{}c|@{}c}
        \hline
        Method & \rotatebox{0}{\textbf{mAP}} & \rotatebox{60}{plane} & \rotatebox{60}{bicycle}	& \rotatebox{60}{bird} & \rotatebox{60}{boat} & \rotatebox{60}{bottle} & \rotatebox{60}{bus} & \rotatebox{60}{car} & \rotatebox{60}{cat} & \rotatebox{60}{chair} & \rotatebox{60}{cow}	& \rotatebox{60}{table} & \rotatebox{60}{dog}	& \rotatebox{60}{horse}	& \rotatebox{60}{bike} & \rotatebox{60}{person} & \rotatebox{60}{plant} & \rotatebox{60}{sheep} & \rotatebox{60}{sofa} & \rotatebox{60}{train} & \rotatebox{60}{tv} \\
        \hline \hline
        
        F-RCNN\cite{ren2015faster}& 73.8 & 86.5 & 81.6 & 77.2 & 58.0 & 51.0 & 78.6 & 76.6 & 93.2 & 48.6 & 80.4 & 59.0 & 92.1 & 85.3 & 84.8 & 80.7 & 48.1 & 77.3 & 66.5 & 84.7 & 65.6 \\
        
        
        SSD513\cite{liu2016ssd} & 79.4 & \underline{90.7} & 87.3 & 78.3 & 66.3 & 56.5 & 84.1 & 83.7 & \textbf{94.2} & 62.9 & 84.5 & 66.3 & \textbf{92.9} & 88.6 & \underline{87.9} & 85.7 & 55.1 & 83.6 & \underline{74.3} & 88.2 & 76.8 \\ 
        
        
        YOLOv2\cite{redmon2017yolo9000} & 73.4 & 86.3 & 82.0 & 74.8 & 59.2 & 51.8 & 79.8 & 76.5 & 90.6 & 52.1 & 78.2 & 58.5 & 89.3 & 82.5 & 83.4 & 81.3 & 49.1 & 77.2 & 62.4 & 83.8 & 68.7 \\
        
        Retina\cite{lin2017focal} & 67.7 & 80.4 & 74.0 & 73.4 & 53.5 & 49.7 & 73.0 & 71.2 & 88.2 & 45.8 & 69.7 & 50.6 & 87.1 & 74.0 & 76.8 & 78.9 & 45.6 & 69.1 & 51.3 & 77.2 & 65.0 \\
        
        ASSD513\cite{yi2019assd} & \underline{81.3} & \textbf{92.1} & \underline{89.2} & \textbf{82.5} & \underline{71.5} & \underline{60.4} & \underline{85.5} & \underline{84.8} & \underline{93.9} & \underline{63.7} & \textbf{88.6} & \underline{67.4} & \underline{92.6} & \underline{90.2} & \textbf{89.0} & \underline{86.5} & \textbf{60.4} & \textbf{88.2} & 73.4 & \underline{88.6} & \underline{77.0} \\
        
        ObjectBox & \textbf{83.7} & \textbf{92.1} & \textbf{92.2} & \underline{80.5} & \textbf{74.0} & \textbf{77.4} & \textbf{92.1} & \textbf{93.3} & 89.5 & \textbf{68.2} & \underline{85.3} & \textbf{75.7} & 87.3 & \textbf{90.6} & 86.6 & \textbf{87.9} & \underline{60.0} & \underline{84.7} & \textbf{77.6} & \textbf{91.3} & \textbf{85.4} \\
        
        \hline
    \end{tabular}
    }
    
    \label{tab:pascal_results}
\end{table*}

\section{Multiscale Prediction} \label{sec:multiscale}
We investigate correlations between the object size and the scale at which it is detected. As shown in Table~\ref{tab:multiscale}, we consider predictions per individual scales, observing that larger objects are better detected at coarser scales, and smaller objects are better detected at finer scales, despite being trained without the bias in defining the positive samples. Each scale level still contributes to the prediction of objects at other scales. This shows that the learning can be better because all objects are being learned at all possible scales.

\begin{table}[h] 
\vspace{-1\baselineskip}
\caption{Predictions per scale level on the MS-COCO dataset}
\centering    
    \begin{tabular}{lcccccccccc}
        \hline
        & \multicolumn{3}{c}{\underline{Avg. Precision, IoU}} & \multicolumn{3}{c}{\underline{Avg. Precision, Area}} & \multicolumn{3}{c}{\underline{Avg. Recall, Area}} \\ 
        Scale level & 
        AP & $AP_{50}$ & $AP_{75}$ & $AP_S$ & $AP_M$ & $AP_L$ & $AR_{S}$ & $AR_{M}$ & $AR_{L}$ \\
        \hline \hline
        
        0, \hspace{1pt} s=\{8\} & 
        {24.9} & {44.9} & 32.1 & {25.5} & {11.7} & {20.5} & {38.1} & {18.4} & {37.3} \\
        
        1, \hspace{1pt} s=\{16\} & 
        {35.7} & {56.4} & {37.7} & {24.3} & {49.2} & {33.6} & {37.6} & {64.8} & {46.3} \\
        
        2, \hspace{1pt} s=\{32\} & 
        {42.7} & {61.6} & {46.0} & {22.0} & {48.0} & {56.1} & {30.9} & {63.2} & {75.6} \\
        
        all, s=\{8,16,32\} & 
        {46.8} & {65.9} & {49.5} & {26.8} & {49.5} & {57.6} & {39.4} & {65.2} & {77.0} \\

        \hline
    \end{tabular}
    \label{tab:multiscale}
\end{table}

\section{IoU Loss}
\label{sec:iou_loss}

In this work, we propose an IoU-based loss
tailored for our object detection method. Recall that our loss is applied to our regression targets $\{L,T,R,B\}$, which are distance values from the corners of the object center cells to the four sides of the bounding box. There are other methods that use similar targets for box regression. For example, FCOS~\cite{tian2019fcos} defines $\{l,t,r,b\}$ as regression targets as distances from each positive location to the bounding box boundaries. The positive locations are selected based on the scale ranges defined for each pyramid level. 
ATSS~\cite{zhang2020bridging} uses the same targets but with a different strategy for positive sample selection. It specifically uses statistical characteristics of objects as the IoU threshold to adaptively select enough positives for each object from appropriate pyramid levels. 

Regardless of their sample selection strategies, both FCOS and ATSS use IoU-based losses, such as the GIoU loss function, for bounding box regression. In our experiments, we replaced their regression losses with our tailored IoU loss, and trained their models with the same settings as the original ones. The results are reported in Table~\ref{tab:iou_losses}. It can be seen that our loss consistently improves detection performance. Our loss improves FCOS by $+0.2\%$ on $AP$, $+0.6\%$ on $AP_{50}$, $+0.1$ on $AP_S$, $+0.9$ on $AP_M$, and $+1.2$ on $AP_L$. Similarly, it achieves a higher performance in ATSS by $+0.4\%$ on $AP$, $+1.1\%$ on $AP_{50}$, $+0.1\%$ on $AP_{75}$, $+0.1$ on $AP_S$, $+1.4$ on $AP_M$, and $+0.6$ on $AP_L$. Note that our loss function is directly applied to the network outputs. It therefore keeps the box integrity based on the model regression targets, and scores the overlapping areas in all four directions.

In Sec. 4.3, we showed the effectiveness of our loss function for box regression, where replacing it with other losses drastically decreased  performance. It was however coupled with the impact of regression location from only one center location. To further investigate the necessity of this loss in our method, we keep the best settings from our ablation study in Sec. 4.3, and only replace our loss with the GIoU loss as used in FCOS and ATSS. Particularly, we use our proposed regression targets ($\{L,T,R,B\}$), augmented centers (the best results in Table 3 part A), one prediction per scale level, and no scale range constraints. As shown in Table~\ref{tab:iou_losses}, ObjectBox with a ResNet-101 backbone clearly benefits from the new IoU loss since it obtains a higher performance by $+1.2\%$ on $AP$, $+1.4\%$ on $AP_{50}$, $+1.0\%$ on $AP_{75}$, $+0.4$ on $AP_S$, $+0.2$ on $AP_M$, and $+1.4$ on $AP_L$, when compared with GIoU loss. The performance boost on ObjectBox with a CSPDarknet backbone is also evident as our loss improves the performance by $+1.1\%$ on $AP$, $+1.7\%$ on $AP_{50}$, $+1.5$ on $AP_{75}$, $+0.7$ on $AP_S$, $+0.6$ on $AP_M$, and $+0.6$ on $AP_L$. These relative improvements indicate that the box IoU loss in our method practically helps to align the bounding boxes more precisely.

\begin{table*}[!t] 
\caption{\small Relative performance of our loss function and GIoU loss, when applied to ObjectBox, FCOS and ATSS }
    \centering
    \begin{tabular}{lclcccccc}
        \hline
         & & & \multicolumn{3}{c}{\underline{Avg. Precision, IoU}} & \multicolumn{3}{c}{\underline{Avg. Precision, Area}}  \\ 

        Method & Backbone & Loss & AP & $AP_{50}$ & $AP_{75}$ & $AP_S$ & $AP_M$ & $AP_L$ \\
        
        \hline \hline
        
        \multirow{2}{*}{FCOS~\cite{tian2019fcos}}  
        &  
         \multirow{2}{*}{ResNeXt-101} & GIoU & 42.1 & 62.1 & 45.2 & 25.6 & 44.9 & 52.0 \\
        
         & 
         & ours & 42.3 & 62.7 & 45.2 & 25.7 & 45.8 & 53.2 \\ \hline
        
        \multirow{2}{*}{ATSS~\cite{zhang2020bridging}} 
         & 
         \multirow{2}{*}{ResNet-101} & GIoU & 43.6 & 62.1 & 47.4 & 26.1 & 47.0 & 53.6 \\
        
         & 
         & ours & 44.0 & 63.2 & 47.5 & 26.2 & 48.4 & 54.2 \\ \hline
        
        \multirow{4}{*}{ObjectBox} 
         & 
         \multirow{2}{*}{ResNet-101} & GIoU & 44.9 & 63.6 & 47.3 & 25.6 & 48.5 & 55.9  \\
        
         & 
         & ours & 46.1 & 65.0 & 48.3 & 26.0 & 48.7 & 57.3  \\ \cline{2-9}
        
         & 
         \multirow{2}{*}{CSPDarknet} & GIoU & 45.7 & 64.2 & 48.0 & 26.1 & 48.9 & 57.0  \\
        
         & 
         & ours & 46.8 & 65.9 & 49.5 & 26.8 & 49.5 & 57.6  \\
        
        \hline
    \end{tabular}
    \label{tab:iou_losses}
\end{table*}

\section{Inference}
\label{sec:inference}

Our method does not impose any additional costs to the inference stage. Given an input image, ObjetcBox predicts an objectness (confidence) score, $m$ classification scores, and four regression values ($\{L,T,R,B\}$) for each feature map location, where $m$ denotes the number of class labels. Therefore, the network output is of size $3\times \frac{W}{s_i} \times \frac{H}{s_i} \times (m+5)$, where $s_i \in \{8,16,32\}$. 
The predictions at all scale levels are sorted based on their confidence scores. They are then refined sequentially based on a threshold value (we set it as $0.001$) until all candidates are investigated. To reduce redundancy in the box prediction, we use non-maximum suppression (NMS)~\cite{neubeck2006efficient} on the predicted boxes based on their classification error. We use an NMS threshold 0.6 and obtain the final results by keeping the highest quality bounding boxes and eliminating the others.

We used the same sizes of input images as in training, and evaluated the inference speed on a single Titan RTX GPU by measuring the end-to-end inference time. We selected different anchor-based and anchor-free methods
as comparisons, including SSD513~\cite{liu2016ssd}, Faster R-CNN with FPN~\cite{lin2017feature}, YOLOv3~\cite{redmon2018yolov3}, FCOS~\cite{tian2019fcos}, and ATSS~\cite{zhang2020bridging}. As shown in Table~\ref{tab:inference}, the average inference speed of ObjectBox with the ResNet-101 backbone is 70 FPS. Meanwhile, using the CSPDarknet backbone can improve the inference speed by 50 FPS, achieving 120 FPS. ObjectBox is significantly faster than other detectors, while having a larger number of parameters (86 M). For instance, the detection speed of ObjectBox is more than two times higher than that of FCOS (50 FPS). Even with the same ResNet-101 backbone, ObjectBox outperforms the ATSS frame rate by $40\%$, i.e. from 50 FPS for ATSS to 70 FPS for ObjectBox. 

The superior time performance of ObjectBox is mainly due to its smaller detection head, and that it considesrs only object central locations for box regression. More specifically, the number of predictions per scale level is just one in ObjectBox, while it is equal to the number of anchors (usually $>1$) in anchor-based detectors. It also filters out all non-center locations by using the confidence score, which undoubtedly reduces the NMS computational load. 
By relaxing the scale range constraints, ObjectBox redefines positive and negative training samples without incurring any additional overheads. It is therefore quite efficient, while achieving the state-of-the-art performance.

\begin{table*}[!t] 
\caption{\small Inference speed comparison }
    \centering
    \begin{tabular}{llccc}
        \hline
        Method & Backbone & \# params & FPS & AP \\
        \hline \hline
        
        SSD513~\cite{liu2016ssd} & ResNet-101 & 57 M & 43 & 31.2 \\
        
        Faster R-CNN w/ FPN~\cite{lin2017feature} & ResNet-101 & 42 M & 26 & 36.2 \\
        
        YOLOv3~\cite{redmon2018yolov3} & DarkNet-53 & 65 M & 20 & 33.0 \\
        
        FCOS~\cite{tian2019fcos} & ResNeXt-101 & 32 M & 50 & 42.1 \\
        
        ATSS~\cite{zhang2020bridging} & ResNet-101 & 32 M & 50 & 43.6 \\
        
        ObjectBox & ResNet-101 & 30 M & 70 & 46.1 \\
        
        ObjectBox & CSPDarknet & 86 M & {\bf 120} & {\bf 46.8} \\
        
        \hline
    \end{tabular}
    \label{tab:inference}
\end{table*}

\section{Qualitative results}
\label{sec:visualization}

\begin{figure}[t] 
\begin{center}
 \includegraphics[width=1\linewidth]{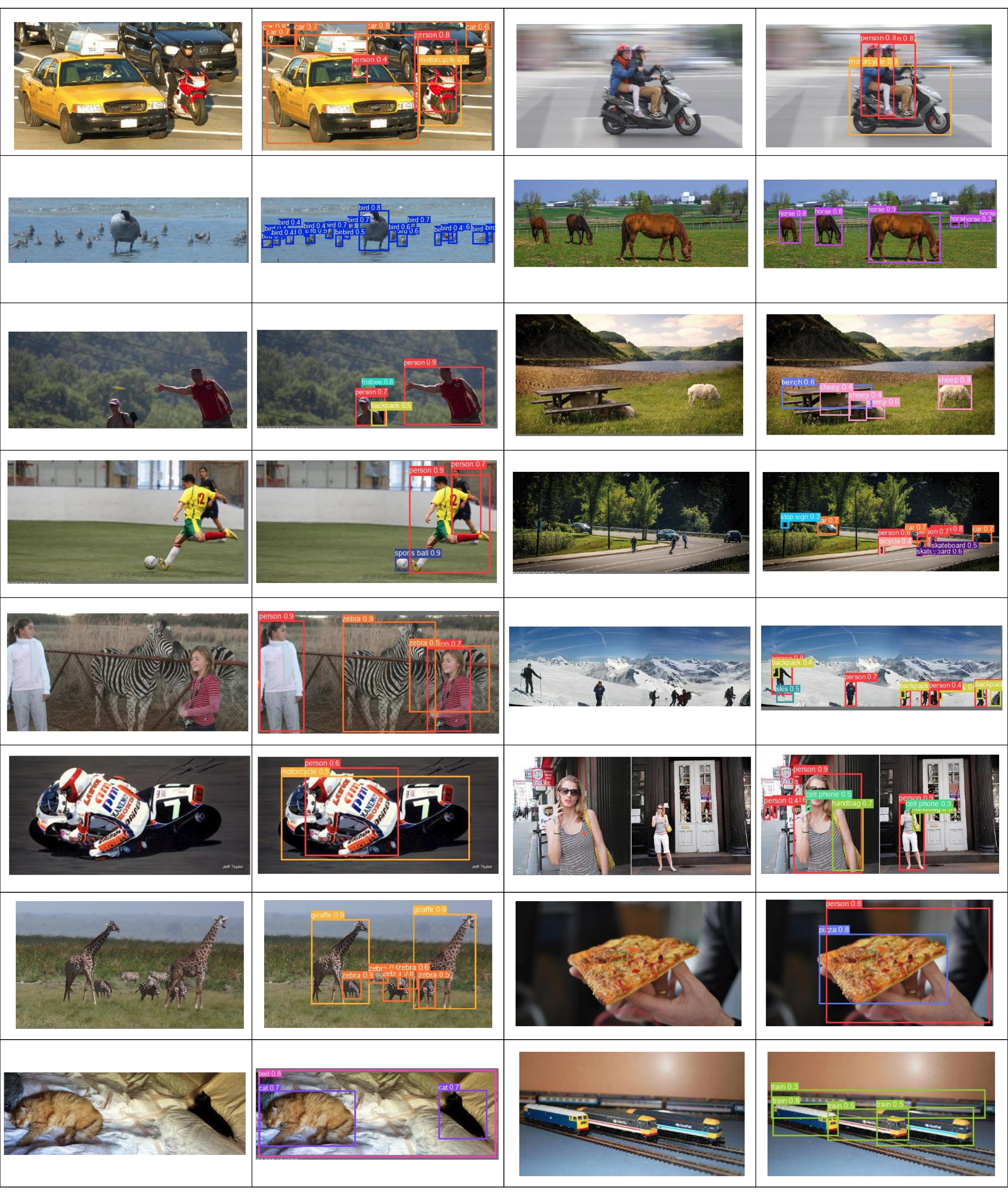}
\end{center}
   \caption{\small Detection examples of applying ObjectBox on COCO test-dev}
   \label{fig:results}
\end{figure}

In Figure~\ref{fig:results}, we show some detection examples on the MS-COCO test-dev dataset. It can be seen that our method is able to successfully detect objects with different sizes and different scene types, with severely overlapping boxes.

\clearpage
%
%
\bibliographystyle{splncs04}
\bibliography{egbib}